\documentclass{article}

\usepackage{arxiv}

\usepackage[utf8]{inputenc} 
\usepackage[T1]{fontenc}    
\usepackage{hyperref}       
\usepackage{url}            
\usepackage{booktabs}       
\usepackage{amsfonts}       
\usepackage{nicefrac}       
\usepackage{microtype}      
\usepackage{lipsum}
\usepackage{times}
\usepackage{epsfig}
\usepackage{graphicx}
\usepackage{amsmath}
\usepackage{amssymb}
\usepackage[ruled]{algorithm2e}
\usepackage{multirow}
\usepackage{amsthm}
\theoremstyle{plain}
\newtheorem*{lft*}{Label Feature Theorem}
\usepackage{bm}

\title{GATCluster: Self-Supervised Gaussian-Attention Network for Image Clustering}

\author{
  Chuang Niu\\
  Xidian University\\
  \texttt{niuchuang@stu.xidian.edu.cn}\\
   \And
  Jun Zhang \\
  Tencent AI Healthcare\\
  \texttt{junejzhang@tencent.com} \\
  \And
  Ge Wang \\
  Rensselaer Polytechnic Institute\\
  \texttt{wangg6@rpi.edu} \\
  \And
  Jimin Liang \\
  Xidian University\\
  \texttt{jimleung@mail.xidian.edu.cn} \\
}

\begin{document}
\maketitle

\date{\vspace{-5ex}}

\begin{abstract}
We propose a self-supervised Gaussian ATtention network for image Clustering (GATCluster). Rather than extracting intermediate features first and then performing the traditional clustering algorithm, GATCluster directly outputs semantic cluster labels without further post-processing. Theoretically, we give a Label Feature Theorem to guarantee the learned features are one-hot encoded vectors and the trivial solutions are avoided. To train the GATCluster in a completely unsupervised manner, we design four self-learning tasks with the constraints of transformation invariance, separability maximization, entropy analysis, and attention mapping. Specifically, the transformation invariance and separability maximization tasks learn the relationships between sample pairs. The entropy analysis task aims to avoid trivial solutions. To capture the object-oriented semantics, we design a self-supervised attention mechanism that includes a parameterized attention module and a soft-attention loss. All the guiding signals for clustering are self-generated during the training process. Moreover, we develop a two-step learning algorithm that is memory-efficient for clustering large-size images. Extensive experiments demonstrate the superiority of our proposed method in comparison with the state-of-the-art image clustering benchmarks. Our code has been made publicly available at \url{https://github.com/niuchuangnn/GATCluster}.
\end{abstract}

\section{Introduction}
\label{sec_intro}

Clustering is the process of separating data into groups according to sample similarity, which is a fundamental unsupervised learning task with numerous applications. Similarity or discrepancy measurement between samples plays a critical role in data clustering. Specifically, the similarity or discrepancy is determined by both data representation and distance function.

Before the extensive application of deep learning, handcrafted features, such as SIFT \cite{SIFT1999} and HoG \cite{HoG2005}, and domain-specific distance functions are often used to measure the similarity. Based on the similarity measurement, various rules were developed for clustering. These include space-partition based (e.g., k-means \cite{kmeans1967}, spectral clustering \cite{Spectral2002}) and hierarchical methods (e.g., BIRCH \cite{BIRCH1996}). With the development of deep learning techniques, researchers have been dedicated to leverage deep neural networks for joint representation learning and clustering, which is commonly referred to as deep clustering. Although significant advances have been witnessed, deep clustering still suffers from an inferior performance for natural images (e.g., ImageNet \cite{ImageNet2015}) in comparison with that for simple handwritten digits in MNIST.

\begin{figure}[t]
\centering
\includegraphics[width=0.8\textwidth]{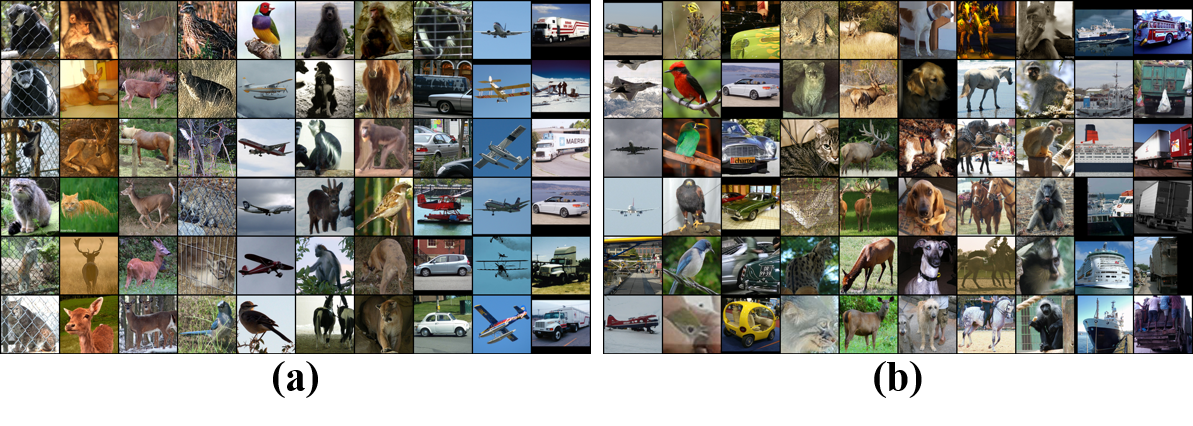}
\caption{Clustering results on STL10. Each column represents a cluster. (a) Sample images clustered by the proposed model without attention, where the clustering principles focus on trivial cues, such as texture (first column), color (second column), or background (fifth column); and (b) Sample images clustered by the proposed model with attention, where the object concepts are well captured.}
\label{fig:cluster_results}
\end{figure}

Various challenges arise when applying deep clustering on natural images.
\emph{First}, many deep clustering methods use stacked auto-encoders (SAE) \cite{Bengio2007Greedy} to extract clustering-friendly intermediate features by imposing some constraints on the hidden layer and the output layer respectively. However, pixel-level reconstruction is not an effective constraint for extracting discriminative semantic features of natural images, since these images usually contain much more instance-specific details that are unrelated to semantics. Recent progresses \cite{DAIC2017}\cite{ADC2019}\cite{Wu_2019_ICCV}\cite{IIC2019} have demonstrated that it is an effective way to directly map data to label features just as in the supervised classification task. However, training such a model in an unsupervised manner is difficult to extract clustering-related discriminative features.
\emph{Second}, clusters are expected to be defined by appropriate semantics (i.e., object-oriented concepts) while current methods tend to group the images by alternative principles (such as color, textures, or background), as shown in Fig. \ref{fig:cluster_results}.
\emph{Third}, the dynamic change between different clustering principles during the training process tend to make the model unstable and easily get trapped at trivial solutions that assign all samples to a single or very few clusters.
\emph{Fourth}, the existing methods were usually evaluated on small images ($32\times32$ to $96\times96$). This is mainly due to the large batch of samples required for training the deep clustering model preventing us from processing large images, especially on memory-limited devices.

To tackle these problems, we propose a self-supervised Gaussian attention network for clustering (GATCluster) that directly outputs discriminative semantic label features. Theoretically, We give a Label Feature Theorem, ensuring that the learned features are one-hot encoded vectors and the trivial solutions can be avoided. To train the GATCluster in a completely unsupervised manner, we design four learning tasks with the constraints of transformation invariance, separability maximization, entropy analysis, and attention mapping. All the guiding signals for clustering are self-generated in the training process.
1) The transformation invariance maximizes the similarity between a sample and its random transformations.
2) The separability maximization task explores both similarity and discrepancy of each paired samples to guide the model learning.
3) The entropy analysis task helps avoid trivial solutions. Different from the nonnegative and $l_2$ norm constraints imposed on label features in \cite{DAIC2017}, we impose the $l_1$ constraint with a probability interpretation. Based on the probability constraint, samples are constrained to be evenly separated by maximizing the entropy, and thus the trivial solutions are avoided.
4) To capture object-orientated semantics, an attention mechanism is proposed based on the observation that the discriminative information on objects is usually presented on local regions. We design a parameterized attention module with a Gaussian kernel and a soft-attention loss that is highly sensitive to discriminative local regions.

For processing large-size images, we develop an efficient two-step learning strategy.
First, the pseudo-targets over a large batch of samples are computed statistically in a split-and-merge manner.
Second, the model is trained on the same batch in a supervised learning manner using the pseudo-targets and in a mini-batch way.
It should be noted that GATCluster is trained by optimizing all loss functions simultaneously instead of alternately.
Our learning algorithm is memory-efficient and thus easy to process large images.

To summarize, the contributions of this paper include

(1) We propose a self-supervised attention network that is trained with four self-learning tasks in a completely unsupervised manner. The data can be partitioned into clusters directly according to semantic label features without further processing during inference.

(2) We give a Label Feature Theorem that can guarantee that the learned features are one-hot encoded vectors and can avoid trivial solutions.

(3) We propose a parameterized attention module with a Gaussian kernel and a soft-attention loss to capture object concepts. To our best knowledge, this is the first attempt in exploring the attention mechanism for unsupervised learning.

(4) Our efficient learning algorithm makes it possible to perform the clustering on large-size images.

(5) Extensive experimental results demonstrate that the proposed GATCluster significantly outperforms or is comparable to the state-of-the-art methods on image clustering datasets.

\section{Related work}
\subsection{Deep Clustering}

We divide the deep clustering methods into two categories: 1) intermediate-feature-based deep clustering and 2) semantic deep clustering. The first category extracts intermediate features and then conducts conventional clustering. The second one directly constructs a nonlinear mapping between original data and semantic label features. By doing so, the samples are clustered just as in the supervised classification task, without any need for additional processing.

Some intermediate-feature-based deep clustering methods usually employ SAE \cite{Hinton2006}\cite{Bengio2007Greedy} or its variants \cite{SDAE2010}\cite{CAE2011}\cite{AAE2015}\cite{VAE2013} to extract intermediate features, and then conduct k-means \cite{DEN2014}\cite{DMC2017} or spectral clustering \cite{DSCN2017}. Instead of  performing representation learning and clustering separately, some studies integrate the two stages into a unified framework \cite{Xie2016}\cite{LI2018161}\cite{DCN2016}\cite{DeepCluster2017}\cite{Zhang_2019_CVPR}\cite{DEPICT2017}\cite{VaDE2017}\cite{GMVAE}\cite{DASC2018}. However, as applied to real scene image datasets, the reconstruction loss of SAE tends to overestimate the importance of low-level features.
In constrast to the SAE-based methods, some methods \cite{Yang2016Joint}\cite{pmlr-v70-hu17b}\cite{CCNN2017} directly use the convolutional neural network (CNN) or multi-layer perceptron (MLP) for representation learning by designing specific loss functions. Unfortunately, the high-dimensional nature of intermediate features are too abundant to effectively reveal the discriminative semantic information of natural images.

Semantic deep clustering methods have recently shown a great promise for clustering. To train such models in the unsupervised manner, various rules have been designed for supervision.
DAC \cite{DAIC2017} recasts clustering into a binary pairwise-classification problem, and the supervised labels are adaptively generated by thresholding the similarity matrix based on label features. It has been theoretically proved that the learned label features are one-hot vectors ideally, and each bit corresponds to a semantic cluster.
As an extension to DAC, DCCM \cite{Wu_2019_ICCV} investigates both pair-wise sample relations and triplet mutual information between deep and shallow layers. However, these two methods are practically susceptible to trivial solutions.
IIC \cite{IIC2019} directly trains a classification network by maximizing the mutual information between original data and their transformations. However, the computation of mutual information requires a very large batch size in the training process, which is challenging to apply on large images.

\subsection{Self-supervised learning}
Self-supervised learning can learn general features by optimizing cleverly designed objective functions of some pretext tasks, in which all supervised pseudo labels are automatically generated from the input data without manual annotations. Various pretext tasks were proposed, including image completion \cite{inpain2016}, image colorization \cite{colorization2016}, jigsaw puzzle \cite{jigsaw2016}, counting \cite{count2017}, rotation \cite{rotation2018}, clustering \cite{vf2018}\cite{Zhang_2019_CVPR}, etc. For the pretext task of clustering, cluster assignments are often used as pseudo labels, which can be obtained by k-means or spectral clustering algorithms.
In this work, both the self-generated relationship of paired samples and object attention are used as the guiding signals for clustering.

\subsection{Attention}

In recent years, the attention mechanism has been successfully applied to various tasks in machine learning and computer vision, such as machine translation \cite{NIPS2017}, image captioning and visual question answering \cite{Anderson2017Bottom}, GAN \cite{Zhang}, person re-identification \cite{Li2018}, visual tracking \cite{Wang2018}, crowd counting \cite{Liu2018}, weakly- and semi-supervised semantic segmentation \cite{Li2018Tell}, and text detection and recognition \cite{He2018}. Given the ground-truth labels, the attention weights are learned to scale-up more related local features for better predictions. However, it is still not explored for deep clustering models that are trained without human-annotated labels.
In this work, we design a Gaussian-kernel-based attention module and a soft-attention loss for learning the attention weights in a self-supervised manner.

\subsection{Learning algorithm of deep clustering}

Various of learning algorithms are designed for training deep clustering models.
Most existing deep clustering models are alternatively trained between updating cluster assignments and network parameters \cite{Xie2016}, or between different clustering heads \cite{IIC2019}. Some of them need pre-training in an unsupervised \cite{Xie2016}\cite{DCN2016}\cite{Zhang_2019_CVPR} or supervised manner \cite{pmlr-v70-hu17b}\cite{CCNN2017}.
On the other hand, some studies \cite{DAIC2017}\cite{Wu_2019_ICCV} directly train the deep clustering models by optimizing all component objective functions simultaneously. However, they do not consider the statistical constraint and are susceptible to trivial solutions.
In this work, we propose a two-step self-supervised learning algorithm that is memory-efficient for processing the large batch training with large-size images.

\section{Method}

\subsection{Label Feature Theorem and problem formulation}
\label{sec_label}

Given a set of samples $\mathcal{X}=\{x_i\}_{i=1}^N$ and the predefined cluster number $k$, the aim of this work is to automatically divide $\mathcal{X}$ into $k$ groups by predicting the label feature $l_i \in R^k$ of each sample $x_i$, where $N$ is the total number of samples.

We first review the Theorem introduced by DAC \cite{Xie2016}. Clustering can be recast as a binary classification problem that measures the similarity and discrepancy between two samples and then determines whether they belong to the same cluster. For each sample $x_i$, the label feature $l_i = f(x_i;w)$ is computed, where $f(\cdot, w)$ is a mapping function with parameters $w$. The parameters $w$ are obtained by minimizing the following objective function:
\begin{equation}
\label{eq_opt1}
\begin{split}
&\min_{w} \textbf{E}(w) = \sum_{i,j}^{N} L(r_{ij}, l_i \cdot l_j), \\
&s.t.\ \forall i \  \|l_i\|_2 = 1,  l_{ih} \ge 0, h = 1, \cdots, k,
\end{split}
\end{equation}
where $r_{ij}$ is the ground-truth relation between sample $x_i$ and $x_j$, i.e., $r_{ij}=1$ indicates that $x_i$ and $x_j$ belong to the same cluster and $r_{ij}=0$ otherwise. In the unsupervised setting, $r_{ij}$ can be estimated by thresholding \cite{DAIC2017}\cite{Wu_2019_ICCV} or clustering introduced in Section \ref{sec_rel}; the inner product $l_i \cdot l_j$ is the cosine distance between two samples due to the label feature is constrained as $\|l_i\|_2 = 1$; $L$ is a loss function instantiated by the binary cross entropy; and $k$ is the predefined cluster number. The theorem proved in \cite{DAIC2017} claimed that if the optimal value of  Eq. \ref{eq_opt1} is attained, the learned label features will be $k$ diverse one-hot vectors. Thus, the cluster identification $c_i$ of image $x_i$ can be directly obtained by selecting the maximum of label features, i.e., $c_i = \mathop{argmax}_{h} \ l_{ih}$. However, it practically tends to obtain trivial solutions that assign all samples to a single or a few clusters. In Appendix \ref{app_dac}, we give a theoretical proof on why it will get trapped at the  trivial solutions to optimize Eq.\ref{eq_opt1} by challenging the Theorem introduced in DAC \cite{DAIC2017}.

Based on above analysis, we formulate clustering as the following optimization problem with a probability constraint and a nonempty cluster constraint:
\begin{equation}
\label{eq_opt2}
\begin{split}
\min_{w} & \textbf{E}(w) = \sum_{i,j}^{N} L(r_{ij}, \frac{l_i}{\|l_i\|_2} \cdot \frac{l_j}{\|l_j\|_2} ) - \sum_{i=1}^N l_i \cdot l_i, \\
s.t.\  &\forall i  \  \|l_i\|_1 = 1, 0 \le  l_{ih} \le 1, h = 1, \cdots, k.  (probability)\\
 &\forall h \  p_h > 0, p_h = \frac{1}{N} \sum_{i=1}^{N} l_{ih}. (nonempty \ cluster)
\end{split}
\end{equation}
For the probability constraint, it is equivalent to impose an extra $L1$-normalization and $ 0 \le  l_{ih} \le 1$ constraints before the $L2$-normalization in Eq. \ref{eq_opt1}. In the nonempty cluster constraint, $p_h$ denotes the frequency of assigning $N$ samples into the $h^{th}$ cluster. And we have a  Label Feature Theorem (the proof of this theorem can be found in Appendix \ref{app_plft}) as following:

\begin{lft*}
\label{theorem}
If the optimal value of Eq. \ref{eq_opt2} is attained, for $\forall i, j, l_i \in E^k, l_i \neq l_j \Leftrightarrow r_{ij}=0$, $l_i = l_j  \Leftrightarrow r_{ij}=1$, and $|\{l_i\}^N_{i=1}| = k$, where $|\cdot|$ denotes the cardinality of a set.
\end{lft*}
The Label Feature Theorem can ensure that the learned features are one-hot encoded vectors in which each bit represents a cluster, and all predefined $k$ clusters are nonempty. However, the learned features may represent various of clustering cues as introduced in Section \ref{sec_intro}. To capture the object-oriented semantics in the unsupervised settings, we propose a Gaussian attention mechanism with a soft-attention loss. Based on the Label Feature Theorem and the attention mechanism, we finally formulate clustering as the following optimization problem,

\begin{equation}
\label{eq_opt3}
\begin{split}
\min_{w} & \textbf{E}(w) = \sum_{i,j=1}^N L_R(r_{ij}, l_i, l_j ) +  \alpha_1 \sum_{i=1}^N L_T(l_i)  + \alpha_2 \sum_{i=1}^N  L_E(l_i) +\alpha_3 \sum_{i=1}^N L_A(l_i, l_i^a), \\
\end{split}
\end{equation}

where $L_R$ and $L_T$ correspond to the first and second item in the objective function of Eq. \ref{eq_opt2},  $L_E$ is to guarantee the nonempty cluster constraint, $L_A$ represents the attention, which is described in Section \ref{sec_att}, and ${\alpha}_1,{\alpha}_2,{\alpha}_3$ are the hyper-parameters to balance the importance of different losses. In practice, the probability constraint can be always true by setting the label features as the outputs of  softmax function. To optimize the problem of Eq. \ref{eq_opt3} for unsupervised clustering, we propose a framework of GATCluster with four self-learning tasks as introduced in the following sections.

\begin{figure*}[ht]
\centering
\includegraphics[width=1.0\textwidth]{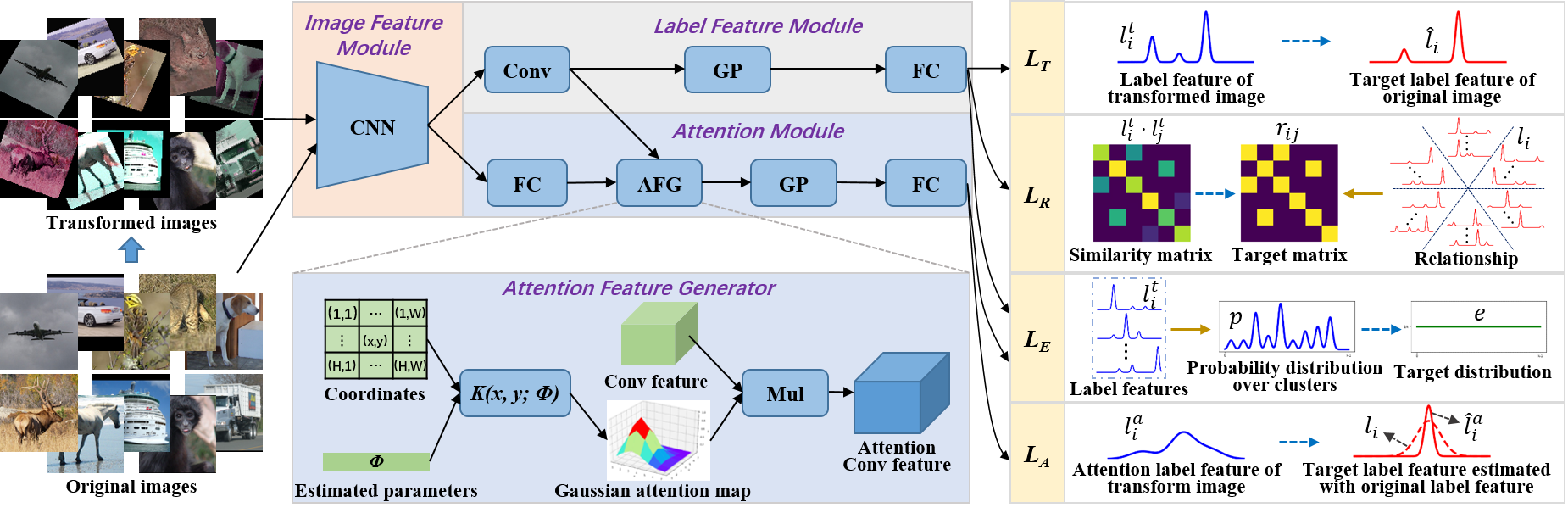}
\caption{GATCluster framework. \emph{CNN} is a convolutional neural network,  \emph{GP} means global pooling, \emph{Mul} represents channel-independent multiplication, \emph{Conv} is a convolution layer, \emph{FC} is a fully connected layer, and \emph{AFG} represents an attention feature generator.}
\label{fig:arc}
\end{figure*}

\subsection{Framework}

GATCluster consists of the following three components: 1) the image feature module, 2) the label feature module, and 3) the attention module, as shown in Figure \ref{fig:arc}. The image feature module extracts image features with a fully convolutional network. The label feature module, which contains a convolutional layer, a global pooling layer and a fully-connected layer, aims to map the image features to label features for clustering. The attention module makes the model focus on discriminative local regions automatically, facilitating the capturing of object-oriented concepts. The attention module consists of three submodules, which are a fully connected layer for estimating the parameters of Gaussian kernel, an attention feature generator, and a global pooling layer followed by another fully connected layer for computing the attention label features. The attention feature generator has three inputs, i.e., the estimated parameter vector $\Phi$, the convolutional feature from the label feature module, and the two-dimensional coordinates of the attention map that are self-generated according to the attention map size $H$ and $W$.

\begin{figure*}[ht]
\centering
\includegraphics[width=0.99\textwidth]{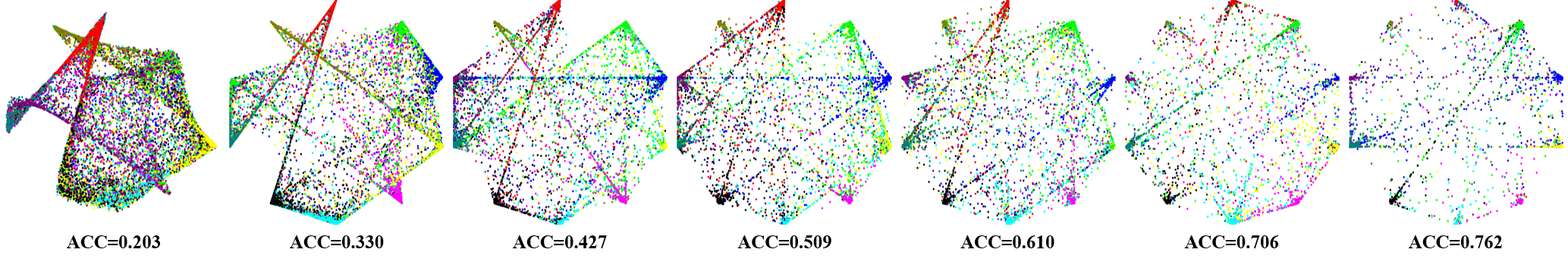}
\caption{Visualization of clustering results in successive training stages (from left to right) for 13K images in ImageNet-10. The results are visualized based on the predicted label features, and each point represents an image and the colors are rendered with the ground-truth label. The corresponding clustering accuracy is presented under each picture. Details can be found in the \ref{app_vis}}
\label{fig:cluster_process}
\end{figure*}

In the training stage, we design four learning tasks driven by transformation invariance, separability maximization, entropy analysis and attention mapping. Specifically, the transformation invariance and separability maximization losses are computed with respect to the predicted label features, the attention loss is estimated with the attention module outputs, and the entropy loss is used to supervise both the label feature module and the attention module. For inference, the image feature module and label feature module are combined as a classifier to suggest the cluster assignments. The clustering results in successive training stages are visualized in Fig. \ref{fig:cluster_process}.

\subsection{Self-learning tasks}

\subsubsection{Transformation invariance task}
\label{sec_trans}

An image after any practically reasonable transformations still reflect the same object. Hence, these transformed images should have same feature representations.
To learn such a similarity, the label feature $l_i$ of original sample $x_i$ is constrained to be close to its transformed counterpart $l_i^t$ of $T(x_i)$, where $T$ is a practically reasonable transformation function. In this work, the transformation function is predefined as the composition of random flipping, random affine transformation, and random color jittering, see Fig. \ref{fig:arc}.
Specifically, the loss function is defined as

\begin{equation}
\label{eq_affine}
L_{T}(l_i^t, \hat{l}_i) = -l_i^t \cdot \hat{l}_i,
\end{equation}
where $\hat{l}_i$ is the target label feature of an original image $x_i$ that is recomputed as:
\begin{equation}
\label{eq_norm}
\hat{l}_{ih} = \frac{l_{ih}/z_h}{\sum_{h'}l_{ih'}/z_{h'}},\ \ z_h=\sum_{j=1}^{M} l_{jh}, h=1, 2, \cdots, k,
\end{equation}
where $M$ is the number of samples, i.e., the batch size used in the training process. Eq. \ref{eq_norm} can balance the sample assignments by dividing the cluster assignment frequency $z_h$, preventing the empty clusters.

\subsubsection{Separability maximization task}
\label{sec_rel}

If the relationships between all pairs of samples are well captured, the label feature will be one-hot encoding vector as introduced in Section \ref{sec_label}. However, the ground-truth relationships cannot be obtained in the unsupervised learning environment. Therefore, we evaluate the relationships of a batch of samples as follows:
\begin{equation}
\label{eq_relation}
r_{ij}=\left\{
\begin{array}{ccc}
1, &    & c_i = c_j \ or \ i=j, \\
0, &    & otherwise,
\end{array}
\right.
\end{equation}
where $c_i = c_j$ indicates that the samples $x_i$ and $x_j$ belong to the same cluster, $i=j$ indicates that the similarity of a sample to itself is 1. To get the cluster id $c_i$, k-means algorithm is conducted on the set of samples based on the predicted label features.

The separability maximization task is to improve the purity of clusters by encouraging samples that are similar to be closer to each other while dissimilar samples to be further away from each other. The loss function is defined as:

\begin{equation}
\label{eq_dis}
\begin{split}
L_{R}(r_{ij}, l_i, l_j) =& -r_{ij}\log(d(l_i, l_j)) - (1-r_{ij})\log(1-d(l_i, l_j)),
\end{split}
\end{equation}
where $d(l_i, l_j) =\frac{l_i}{\|l_i\|_2} \cdot \frac{l_j}{\|l_j\|_2}$ is the cosine distance.

\subsubsection{Entropy analysis task}
\label{sec_ent}

The entropy analysis task is designed to avoid trivial solutions by satisfying the nonempty cluster constraint in Eq. \ref{eq_opt2}.
We maximize the entropy of the empirical probability distribution $p$ over $k$ cluster assignments. Thus, the loss function is defined as

\begin{equation}
\label{eq_entropy}
\begin{split}
L_{E}(l_1, \cdots, l_m) = \sum_{h=1}^{k} p_h \log(p_h), \\
 p_h = \frac{1}{m} \sum_{i=1}^{m} l_{ih}, h = 1, \cdots, k,
 \end{split}
\end{equation}
where $p$ is estimated with the predicted label features of $m$ samples, which can be a subset of the whole batch.
Actually, maximizing the entropy will steer $p$ towards a uniform distribution (denoted by $e$ in Fig. \ref{fig:arc}), thus, the nonempty constraint that $\forall h, p_h = \frac{1}{m} > 0$ is satisfied so that the trivial solutions are avoided as in \textbf{Label Feature Theorem}.

\subsubsection{Attention mapping task}
\label{sec_att}

The attention mapping task aims to make the model recognize the most discriminative local regions concerning the whole image semantic. The basic idea is that the response to the discriminative local regions should be more intense than that to the entire image. To this end, there are two problems to be solved: 1) how to design the attention module for localizing the discriminative local regions? and 2) how to train the attention module in a self-supervised manner?

With regard to the first problem, we design a two-dimensional Gaussian kernel $K(u;\Phi)$ to generate an attention map $A$ as:
\begin{equation}
\label{eq_gaussian}
\begin{split}
A(x, y)& = K(u;\Phi) = e^{-\frac{1}{\alpha}(u-\mu)^T\Sigma^{-1}(u-\mu)}, \\
&x=1, \cdots, H,\ and \ y=1, \cdots, W,
\end{split}
\end{equation}
 where $u=[x, y]^T$ denotes the coordinate vector, $\Phi = [\mu, \Sigma]$ denotes the parameters of the Gaussian kernel, $\mu=[\mu_x, \mu_y]^T$ is the mean vector that defines the most discriminative location, $\Sigma \in \textbf{R}^{2 \times 2}$ is the covariance matrix that defines the shape and size of a local region, $\alpha$ is a predefined hyper parameter, and H and W are the height and width of the attention map. In our implementation, the coordinates are normalized over $[0, 1]$.
 Taking CNN features as input, a fully connected layer is used to estimate the parameter $\Phi$. Then, the model can focus on the discriminative local region by multiplying each channel of convolutional features with the attention map. The weighted features are mapped to the attention label features using a global pooling layer and a fully connected layer, as shown in Fig. \ref{fig:arc}. It should be noted that there are also alternative designs of the attention module to generate attention maps, such as a convolution layer followed by a sigmoid function. However, we obtained better results with the parameterized Gaussian attention module due to that it has a much less number of parameters to be estimated, and the Gaussian attention map fits for capturing the object-oriented concepts of natural images.

 With regard to the second problem, we define a soft-attention loss as
 \begin{equation}
 \label{eq_att}
 \begin{split}
 L_{A}(l_i^a, \hat{l}_i^a) = & \frac{1}{k}\sum_{h=1}^k -\hat{l}_{ih}^a \log(l_{ih}^a)  - (1-\hat{l}_{ih}^a) \log(1-l_{ih}^a),
 \end{split}
 \end{equation}
 \begin{equation}
 \label{eq_att1}
 \hat{l}_{ih}^a = \frac{l_{ih}^2/z_h}{\sum_{h'}l_{ih'}^2/z_{h'}}, h = 1, \cdots, k,
 \end{equation}
 where $l_i^a$ is the output of the attention module, $\hat{l}_i^a$ is the target label feature for regression, and $z_h$ is the same as in Eq. \ref{eq_norm} to balance the cluster assignments. As defined in Eq. (\ref{eq_att1}), the target label feature $\hat{l}_i^a$ encourages the current high scores and suppresses low scores of the whole image label feature $l_i$, thus making $\hat{l}_i^a$ a more confident version of the whole image label feature $l_i$, see Fig. \ref{fig:arc} for demonstration. By doing so, the local image region, which is localized by the attention module, is discriminative in terms of the whole image semantics. In practice, the local region usually presents the expected object semantic as shown in Fig. \ref{fig:examples}.

\begin{algorithm}
\scriptsize
\label{alg_1}
\caption{\scriptsize{GATCluster learning algorithm.}}
\LinesNumbered
\KwIn{Dataset $\mathcal{X}=\{x_i\}_{i=1}^N, k, M$, $ m_1$, $m_2$
}
\KwOut{Cluster label $c_i$ of $x_i \in \mathcal{X}$}
Randomly initialize network parameters \textbf{w}\;
Initialize e = 0\;
\While{$e < $ total epoch number}{

    \For{$b \in \{1, 2, \dots, \lfloor\frac{N}{M}\rfloor\}$}{
            Select $M$ samples as $\mathcal{X}_b$ from $\mathcal{X}$ \;
            \textbf{\emph{Step-1}:} \\
             \For{$u \in \{1, 2, \dots, \lfloor\frac{M}{m_1}\rfloor\}$}{
                  Select $m_1$ samples as $\mathcal{X}_u$ from $\mathcal{X}_b$\;
                  Calculate the label features of $\mathcal{X}_u$\;
             }
            Concatenate all label features of $M$ samples \;
            Calculate pseudo targets $T_b=\{(\hat{l}_i, r_{ij}, \hat{l}_i^a)\}$ of $\mathcal{X}_b$ with Eqs. \ref{eq_norm}, \ref{eq_relation}, and \ref{eq_att1}\;
            \textbf{\emph{Step-2}:} \\
            Randomly transform samples in $\mathcal{X}_b$ as $\mathcal{X}_b^t$ \;
            \For{$v \in \{1, 2, \dots, \lfloor\frac{M}{m_2}\rfloor\}$}{
                Randomly select $m_2$ samples as $[\mathcal{X}_v; T_v]$ from [$\mathcal{X}_b^t$; $T_b$] \;
                Optimize \textbf{w} on $[\mathcal{X}_v; T_v]$ by minimizing Eq. \ref{eq_loss} using Adam \;
            }

    }
    $e := e + 1$
}
\ForEach{$x_i \in \mathcal{X}$}{
        $l_i := f(x_i; \textbf{w})$ \;
        $c_i:=\arg\max_h(l_{ih})$\;
}
\end{algorithm}

\subsection{Learning algorithm}

We develop a two-step learning algorithm that combines all the self-learning tasks to train GATCluster in an unsupervised learning manner. The total loss function is
\begin{equation}
\label{eq_loss}
L_{total} = L_{R} + \alpha_1 L_{T} +\alpha_2 L_{A} + \alpha_3 L_{E}
\end{equation}
where the entropy loss is computed with the label features $l_i$ and $l_i^a$ predicted by the label feature module and attention module respectively, i.e., $L_{E} = L_{E}(l_1,\cdots, l_M) + L_{E}(l_1^a, \cdots, l_M^a)$. $\alpha_i, i=1,2,3$ are hyper parameters to weight the tasks.

The proposed two-step learning algorithm is presented in Algorithm \ref{alg_1}. Since deep clustering methods usually require a large batch of samples for training, it is difficult to apply to large images on a memory-limited device. To tackle this problem, we divide the large-batch-based training process into two steps for each iteration. The first step statistically calculates the pseudo-targets for a large batch of $M$ samples using the model trained in the last iteration. To achieve this with a memory-limited device, we further split a large batch into sub-batches of $m_1$ samples and calculate the label features for each sub-batch independently. Then, all label features of $M$ samples are concatenated for computation of the pseudo labels. The second step trains the model just as in supervised learning with a mini-batch of $m_2$ samples iteratively.

\section{Experiments and Results}

\subsection{Data}
We evaluated our and the compared deep clustering methods on five datasets, including STL10 \cite{STL2011} that contains 13K $96\times96$ images of 10 clusters, ImageNet-10 \cite{DAIC2017} that contains 13K images of 10 clusters, ImageNet-Dog \cite{DAIC2017} that contains 19.5K images of 15 clusters of dogs, Cifar10 and Cifar100-20 \cite{CIFAR2009}. The image size of ImageNet-10 and ImageNet-Dog is around $500\times300$. Cifar10 and Cifar100-20 both contain 60K $32\times32$ images, and have 10 and 20 clusters respectively.

\subsection{Implementation details}
At the training stage, especially at the beginning, samples tend to be clustered by color cues. Therefore, we took grayscale images as inputs except for ImageNet-Dog, as the color plays an important role in differentiating the sub-categories of dogs. It is noted that the images are converted to grayscale after applying the random color jittering during training.
The $L1$-normalization was implemented by a softmax layer.
For simplicity, we assume $\Sigma=\begin{bmatrix} \delta & 0 \\ 0 & \delta \end{bmatrix}$ and, thus, there are only three parameters of Gaussian kernel to be estimated, i.e., $[\mu_x, \mu_y; \delta]$.
We used Adam to optimize the network parameters and the base learning rate was set to 0.001. We set the batch size $M$ to 1000 for STL10 and ImageNet-10, 1500 for ImageNet-Dog, 4000 for Cifar10, and 6000 for Cifar100-20. The sub-batch size $m_1$ in calculating pseudo targets can be adjusted according to the device memory and will not affect the results. The sub-batch size $m_2$ was 32 for all experiments.
Hyper parameters $\alpha$, $\alpha_1$, $\alpha_2$, and $\alpha_3$ were empirically set to 0.05, 5, 5, and 3 respectively.

\subsection{Network architecture}
In all experiments, we used the VGG-style convolutional network with batch normalization to implement the image feature extraction module. The main difference between the architectures for different experiments are the layer number, kernel-size, and the output cluster number. The details of these architectures can be found in the Appendix \ref{app_arc}.

\subsection{Evaluation metrics}
We used three popular metrics to evaluate the performance of the involved clustering methods, including Adjusted Rand Index (ARI) \cite{hubert1985comparing}, Normalized Mutual Information (NMI) \cite{strehl2002clusterensembles} and clustering Accuracy (ACC) \cite{LiD06}.

\begin{table*}[ht]
\footnotesize
\renewcommand{\arraystretch}{1.3}
\renewcommand\tabcolsep{1.5pt}
\caption{ Comparison with the existing methods. GATCluster-128 resizes input images to $128\times128$ for ImageNet-10 and ImageNet-Dog while other models take $96\times96$ images as inputs. On Cifar10 and Cifar100, the input size is $32\times32$. The best three results are highlighted in \textbf{bold} .}
\label{table_results_compare}
\centering
\begin{tabular}{|c|ccc|ccc|ccc|ccc|ccc|}
\hline

\multirow{2}{*}{Method}                            & \multicolumn{3}{c|}{STL10}&\multicolumn{3}{c|}{ImageNet-10}&\multicolumn{3}{c|}{ImageNet-dog}&\multicolumn{3}{c|}{Cifar10}&\multicolumn{3}{c|}{Cifar100-20}\\
                                                                        \cline{2-16}
                                                                        & ACC  & NMI  & ARI& ACC&NMI&ARI   & ACC&NMI&ARI  &ACC&NMI&ARI   &ACC&NMI&ARI\\
\hline\hline
k-means \cite{kmeans1967}                     & 0.192   & 0.125 & 0.061&   0.241 &0.119& 0.057   & 0.105 &0.055&0.020   & 0.229 &0.087&0.049 & 0.130 &0.084&0.028\\
SC \cite{Spectral2002}                             & 0.159   & 0.098 & 0.048&   0.274 &0.151&0.076    & 0.111 &0.038&0.013   & 0.247 &0.103&0.085 & 0.136 &0.090&0.022\\
AC \cite{Pasi2006Fast}                            & 0.332   & 0.239 & 0.140&   0.242 &0.138&0.067    & 0.139 &0.037&0.021   & 0.228 &0.105&0.065 & 0.138 &0.098&0.034\\
NMF \cite{NMF}                                      & 0.180   & 0.096 & 0.046&   0.230 &0.132&0.065    & 0.118 &0.044&0.016   & 0.190 &0.081&0.034 & 0.118 &0.079&0.026\\
AE \cite{Bengio2007Greedy}                   & 0.303   & 0.250 & 0.161&   0.317 &0.210&0.152    & 0.185 &0.104&0.073   & 0.314 &0.239&0.169 &0.165 &0.100&0.048\\
SAE \cite{Bengio2007Greedy}                 & 0.320   & 0.252 & 0.161&   0.335 &0.212&0.174    & 0.183 &0.113&0.073   & 0.297 &0.247&0.156& 0.157 &0.109&0.044\\
SDAE \cite{SDAE2010}                           & 0.302   & 0.224 & 0.152&   0.304 &0.206&0.138    & 0.190 &0.104&0.078   & 0.297 &0.251&0.163& 0.151 &0.111&0.046\\
DeCNN \cite{Zeiler2010Deconvolutional}& 0.299   & 0.227 & 0.162&   0.313 &0.186&0.142    & 0.175 &0.098&0.073   & 0.282 &0.240&0.174&0.133 &0.092&0.038\\
SWWAE \cite{SWWAE2015}                  & 0.270   & 0.196 & 0.136&   0.324 &0.176&0.160    & 0.159 &0.094&0.076   & 0.284 &0.233&0.164&0.147&0.103&0.039\\
CatGAN   \cite{catgan2016}                     & 0.298   & 0.210 & 0.139&   0.346 &0.225&0.157    & N/A &N/A&N/A   & 0.315 &0.265&0.176   & N/A &N/A&N/A\\
GMVAE   \cite{GMVAE}                         & 0.282   & 0.200 & 0.146&   0.334 &0.193&0.168    & N/A &N/A&N/A   & 0.291 &0.245&0.167   & N/A &N/A&N/A\\
JULE-SF \cite{Yang2016Joint}            & 0.274   & 0.175 & 0.162&   0.293 &0.160&0.121    & N/A &N/A&N/A   & 0.264 &0.192&0.136   & N/A &N/A&N/A\\
JULE-RC \cite{Yang2016Joint}           & 0.277   & 0.182 & 0.164&   0.300 &0.175&0.138   & 0.138 &0.054&0.028   & 0.272 &0.192&0.138    & 0.137 &0.103&0.033\\
DEC \cite{Xie2016}                             & 0.359   & 0.276 & 0.186&   0.381 &0.282&0.203    & 0.195 &0.122&0.079   & 0.301 &0.257&0.161   & 0.185 &0.136&0.050\\
DAC$^*$ \cite{DAIC2017}                 & 0.434   & 0.347 & 0.235&   0.503 &0.369&0.284    & 0.246 &0.182&0.095   & 0.498 &0.379&0.280   & 0.219 &0.162&0.078\\
DAC \cite{DAIC2017}                         & 0.470   & \textbf{0.366} & \textbf{0.257}&   0.527 &0.394&0.302   & 0.275 &0.219&0.111   & 0.522 &\textbf{0.396}&\textbf{0.306}   & 0.238 &\textbf{0.185}&\textbf{0.088}\\
IIC  \cite{IIC2019}                      & \textbf{0.499}   & N/A & N/A&  N/A &N/A&N/A      & N/A &N/A&N/A& \textbf{0.617}& N/A & N/A& \textbf{0.257}& N/A & N/A\\
DCCM \cite{Wu_2019_ICCV}  & \textbf{0.482}    & \textbf{0.376} & \textbf{0.262}& \textbf{0.710} & \textbf{0.608} &\textbf{0.555}& \textbf{0.383}&\textbf{0.321} &\textbf{0.182}& \textbf{0.623} & \textbf{0.496}&\textbf{0.408}& \textbf{0.327} &\textbf{0.285}&\textbf{0.173}\\

\hline
GATCluster   & \textbf{0.583}   & \textbf{0.446} & \textbf{0.363}& \textbf{0.739}   & \textbf{0.594} & \textbf{0.552}& \textbf{0.322}& \textbf{0.281} & \textbf{0.163} & \textbf{0.610}& \textbf{0.475} & \textbf{0.402}   & \textbf{0.281} &\textbf{0.215}&\textbf{0.116}\\

GATCluster-128 & N/A   & N/A & N/A& \textbf{0.762}   & \textbf{0.609} & \textbf{0.572}& \textbf{0.333}   & \textbf{0.322} & \textbf{0.200}& N/A   & N/A & N/A& N/A   & N/A & N/A \\
\hline

\end{tabular}
\end{table*}

\subsection{Comparison with existing methods}
Table \ref{table_results_compare} presents a comparison with the existing methods. Under the same conditions, the proposed method significantly improves the clustering performance by 8\%, 7\%, and 10\% approximately compared with the best of the others in terms of ACC, NMI and ARI on STL10. On ImageNet-10, ACC is improved by 5\% compared with the strong baseline that is set by the most recently proposed DCCM \cite{Wu_2019_ICCV}.  On the sub-category dataset ImageNet-Dog, our method achieves results comparable to that of DCCM. Moreover, our method is capable of processing large images, and in that case the clustering results are further improved. On the small image datasets, i.e., Cifar10 and Cifar100-20, the proposed method also achieves comparable performance relative to the state-of-the-art. Importantly, our GATCluster has the interpretability to the learned cluster semantics by presenting the corresponding local regions. The above results strongly demonstrate the superiority of our proposed method.

\begin{figure}[ht]
\centering
\includegraphics[width=0.79\textwidth]{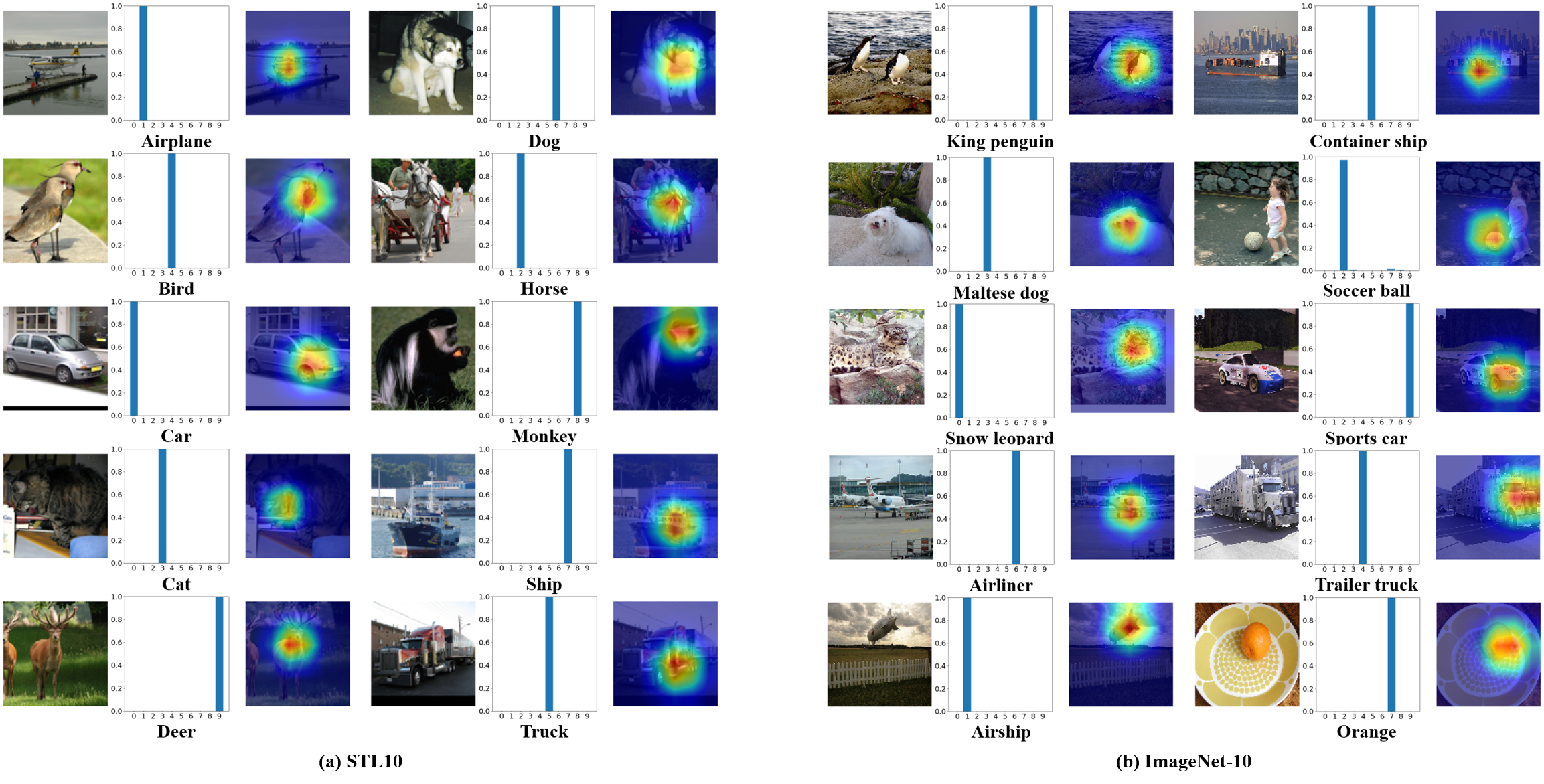}
\caption{Visualization of GATCluster on STL10 and ImageNet10. For each class, the example image, the predicted label features, and the attention map overlaid on the image are shown from left to right.}
\label{fig:examples}
\end{figure}

\subsection{Ablation study}

To validate the effectiveness of each component, we conducted the ablation studies as shown in Table \ref{table_results_abl}. Similar to \cite{ADC2019}, each variant was evaluated ten times and the best accuracy, average accuracy and the standard deviation are reported. Table \ref{table_results_abl} demonstrates that the best accuracy is achieved when all learning tasks are used with grayscale images.
Particularly, the attention module improves the accuracy by up to 4.4 percent for the best accuracy and 4.3 percent for average accuracy. This is attributed to that the attention module has the ability to localize the discriminative regions with respect to the whole image semantic, and thus it can well capture the expected object-oriented semantics, as shown in Figure \ref{fig:examples}. In addition, the color information is a strong distraction for object clustering, and better clustering results can be obtained after the color images are changed to grayscale.

Because it is very easy to get trapped at trivial solutions if the entropy loss is missed in our experiments, we do not show the results from the ablation of the entropy analysis task.

\begin{table}[ht]
\footnotesize
\renewcommand{\arraystretch}{1.3}
\renewcommand\tabcolsep{2pt}
\caption{ Ablation studies of GATCluster on STL10, where TI, SM and AM mean transformation invariance, separability maximization, and attention mappling respectively. Each algorithm variant was evaluated ten times.}
\label{table_results_abl}
\centering
\begin{tabular}{|c|ccc|ccc|ccc|}

\hline

\multirow{2}{*}{Method}                          & \multicolumn{3}{c|}{ACC}&\multicolumn{3}{c|}{NMI}&\multicolumn{3}{c|}{ARI}\\
\cline{2-10}
                                                                        &Best &Mean & Std&Best &Mean & Std&Best &Mean & Std\\
\hline\hline
Color                                                             & 0.556 & 0.517 & 0.034     & 0.427   & 0.402 & 0.022     & 0.341   & 0.298 & 0.031 \\
No TI                                            & 0.576& 0.546 & 0.016     & 0.435   & 0.417 & 0.012     & 0.347   & 0.325 & 0.014 \\
No SM                                                    & 0.579 & 0.529 & 0.029     & 0.438   & 0.412 & 0.019     & 0.356   & 0.310 & 0.024\\
No AM                                                  & 0.539 & 0.494 & 0.020     & 0.416   & 0.383 & 0.015     & 0.316   & 0.282 & 0.013\\
\hline
Full setting                                                    & \textbf{0.583}& 0.537 & 0.033     & \textbf{0.446}& 0.415 &0.022    &\textbf{0.363}&0.315&0.032\\
\hline

\end{tabular}
\end{table}

\subsection{Effectiveness of image size}
  The biggest image size used by most of the existing unsupervised clustering methods is not larger than $96\times96$ (e.g., in STL10). However, images in the modern datasets usually have much larger sizes, which are not effectively explored by unsupervised deep clustering methods. With the proposed two-step learning algorithm, we are able to process large images. An interesting question then arises: will large images help produce a better clustering accuracy? To answer this question, we explored the effect of image size on clustering results. Specifically, we used the ImageNet-10 in our experiments. We evaluated four input image sizes, i.e., $96\times96$, $128\times128$, $160\times160$, and $192\times192$ by simply resizing the original images. For each image size, the model architecture is slightly different as described in the \ref{app_arc}. We conducted five experimental trails for each image size and report the best and average accuracies as well as the standard deviation in Table \ref{table_results_size}. The results show that the clustering performance is significantly improved when the image size is increased from $96\times96$ to $128\times128$. It is demonstrated that taking the larger images as inputs can be beneficial for clustering.

Practically, our proposed methods can be performed on much larger size of images. The clustering results are not further improved when the image size is larger than $128\times128$. It may be due to that networks will become deepened with an increased image size, and thus there is a trade-off between the number of network parameters and the size of the training dataset. However, it is valuable to explore larger size of images for clustering in the future.

\begin{table}[ht]
\footnotesize
\renewcommand{\arraystretch}{1.2}
\renewcommand\tabcolsep{2pt}
\caption{ Clustering results of different image sizes on ImageNet-10. Each setting was evaluated five times.}
\label{table_results_size}
\centering
\begin{tabular}{|c|ccc|ccc|ccc|}
\hline

\multirow{2}{*}{Image size}                      & \multicolumn{3}{c|}{ACC}&\multicolumn{3}{c|}{NMI}&\multicolumn{3}{c|}{ARI}\\
\cline{2-10}
                                                                       &Best &Mean & Std&Best &Mean & Std&Best &Mean & Std\\
\hline\hline
$96\times96$                                                 & 0.739 & 0.708 & 0.031     & 0.594   & 0.581 & 0.012     & 0.552   & 0.529 & 0.019 \\
$128\times128$                                             & \textbf{0.762} & 0.735 & 0.020     & 0.609   & 0.592 & 0.013     & 0.572   & 0.544 & 0.023 \\
$160\times160$                                             & 0.712 & 0.669 & 0.033     & 0.567   & 0.511 & 0.043     & 0.500   & 0.453 & 0.039\\
$192\times192$                                             & 0.738 & 0.608 & 0.067     & 0.612   & 0.474 & 0.071     & 0.559   & 0.405 & 0.079\\
\hline

\end{tabular}
\end{table}

\subsection{Effectiveness of the attention map resolution}
 A high-resolution attention map will provide precise location but weaken the global semantics. We evaluated the effect of the attention map resolution on the clustering results for Image-10. We set the input image size in this experiment to $128\times128$, and evaluate five attention map resolutions (in pixels): $2\times2$, $4\times4$, $6\times6$, $8\times8$, and $10\times10$ as shown in Table \ref{table_attention_size}. It is demonstrated that the attention map resolution of $6\times6$ achieves the best results.

\begin{table}[ht]
\footnotesize
\renewcommand{\arraystretch}{1.2}
\renewcommand\tabcolsep{2pt}
\caption{ Clustering results with different attention map resolutions on ImageNet-10. Each setting was evaluated five times.}
\label{table_attention_size}
\centering
\begin{tabular}{|c|ccc|ccc|ccc|}
\hline
\multirow{2}{*}{Resolution}                      & \multicolumn{3}{c|}{ACC}&\multicolumn{3}{c|}{NMI}&\multicolumn{3}{c|}{ARI}\\
\cline{2-10}
                                                                       &Best &Mean & Std&Best &Mean & Std&Best &Mean & Std\\
\hline\hline
$2\times2$                                                & 0.746 & 0.666 & 0.050     & 0.625   & 0.538 & 0.050     & 0.569   & 0.477 & 0.045 \\
$4\times4$                                                & 0.706 & 0.678 & 0.017     & 0.539   & 0.528 & 0.012     & 0.486   & 0.473 & 0.014 \\
$6\times6$                                                & \textbf{0.762} & 0.735 & 0.020     & 0.609   & 0.592 & 0.013     & 0.571   & 0.544 & 0.023 \\
$8\times8$                                                & 0.742 & 0.719 & 0.018     & 0.618   & 0.594 & 0.019     & 0.561   & 0.536 & 0.018 \\
$10\times10$                                            & 0.671 & 0.645 & 0.020     & 0.549   & 0.520 & 0.021     & 0.478   & 0.450 & 0.020 \\
\hline

\end{tabular}
\end{table}

\section{Conclusion}

For deep unsupervised clustering, we have proposed the GATCluster model to learn discriminative semantic label features with four self-learning tasks. Specifically, the transformation invariance and separability maximization tasks explore the similarity and discrepancy between samples. The designed attention mechanism facilitates the formation of object concepts during the training process. The entropy loss can effectively avoid trivial solutions. With all learning tasks, the developed two-step learning algorithm is both training-friendly and memory-efficient and thus is capable of processing large images. The GATCluster model has a potential for clustering the images with complex contents and the object discovery in the unsupervised settings.

\section{Acknowledgments}
The research was supported by the National Natural Science Foundation of China (61976167, 61571353, U19B2030) and the Science and Technology Projects of Xi’an, China (201809170CX11JC12).

\bibliographystyle{unsrt}
\bibliography{gatcluster}

\begin{thebibliography}{10}

\bibitem{SIFT1999}
David~G. Lowe.
\newblock Lowe, d.: Object recognition from local scale-invariant features. in:
  Proc. iccv.
\newblock In {\em ICCV}, 1999.

\bibitem{HoG2005}
N.~{Dalal} and B.~{Triggs}.
\newblock Histograms of oriented gradients for human detection.
\newblock In {\em CVPR}, volume~1, pages 886--893 vol. 1, June 2005.

\bibitem{kmeans1967}
J.~Macqueen.
\newblock Some methods for classification and analysis of multivariate
  observations.
\newblock In {\em In 5-th Berkeley Symposium on Mathematical Statistics and
  Probability}, pages 281--297, 1967.

\bibitem{Spectral2002}
Andrew~Y. Ng, Michael~I. Jordan, and Yair Weiss.
\newblock On spectral clustering: Analysis and an algorithm.
\newblock In T.~G. Dietterich, S.~Becker, and Z.~Ghahramani, editors, {\em
  NeurIPS}, pages 849--856. MIT Press, 2002.

\bibitem{BIRCH1996}
Tian Zhang, Raghu Ramakrishnan, and Miron Livny.
\newblock Birch: An efficient data clustering method for very large databases.
\newblock In {\em SIGMOD Conference}, 1996.

\bibitem{ImageNet2015}
Olga Russakovsky, Jia Deng, Hao Su, Jonathan Krause, Sanjeev Satheesh, Sean Ma,
  Zhiheng Huang, Andrej Karpathy, Aditya Khosla, and Michael Bernstein.
\newblock Imagenet large scale visual recognition challenge.
\newblock {\em International Journal of Computer Vision}, 115(3):211--252,
  2015.

\bibitem{Bengio2007Greedy}
Y.~Bengio, P.~Lamblin, D.~Popovici, H.~Larochelle, and U.~Montreal.
\newblock Greedy layer-wise training of deep networks.
\newblock {\em NeurIPS}, 19:153--160, 2007.

\bibitem{DAIC2017}
J.~{Chang}, L.~{Wang}, G.~{Meng}, S.~{Xiang}, and C.~{Pan}.
\newblock Deep adaptive image clustering.
\newblock In {\em ICCV}, pages 5880--5888, Oct 2017.

\bibitem{ADC2019}
Philip Haeusser, Johannes Plapp, Vladimir Golkov, Elie Aljalbout, and Daniel
  Cremers.
\newblock Associative deep clustering: Training a classification network with
  no labels.
\newblock In Thomas Brox, Andr{\'e}s Bruhn, and Mario Fritz, editors, {\em
  Pattern Recognition}, pages 18--32, Cham, 2019. Springer International
  Publishing.

\bibitem{Wu_2019_ICCV}
Jianlong Wu, Keyu Long, Fei Wang, Chen Qian, Cheng Li, Zhouchen Lin, and
  Hongbin Zha.
\newblock Deep comprehensive correlation mining for image clustering.
\newblock In {\em ICCV}, October 2019.

\bibitem{IIC2019}
Xu~Ji, Joao~F. Henriques, and Andrea Vedaldi.
\newblock Invariant information clustering for unsupervised image
  classification and segmentation.
\newblock In {\em ICCV}, October 2019.

\bibitem{Hinton2006}
G~E Hinton and R~R Salakhutdinov.
\newblock Reducing the dimensionality of data with neural networks.
\newblock {\em Science}, 313(5786):504--507, July 2006.

\bibitem{SDAE2010}
Pascal Vincent, Hugo Larochelle, Isabelle Lajoie, Yoshua Bengio, and
  Pierre~Antoine Manzagol.
\newblock Stacked denoising autoencoders: Learning useful representations in a
  deep network with a local denoising criterion.
\newblock {\em Journal of Machine Learning Research}, 11(12):3371--3408, 2010.

\bibitem{CAE2011}
Jonathan Masci, Ueli Meier, Dan Cire{\c{s}}an, and J{\"u}rgen Schmidhuber.
\newblock Stacked convolutional auto-encoders for hierarchical feature
  extraction.
\newblock In Timo Honkela, W{\l}odzis{\l}aw Duch, Mark Girolami, and Samuel
  Kaski, editors, {\em Artificial Neural Networks and Machine Learning}, pages
  52--59, Berlin, Heidelberg, 2011. Springer Berlin Heidelberg.

\bibitem{AAE2015}
Alireza Makhzani, Jonathon Shlens, Navdeep Jaitly, and Ian~J. Goodfellow.
\newblock Adversarial autoencoders.
\newblock {\em CoRR}, abs/1511.05644, 2015.

\bibitem{VAE2013}
Diederik~P Kingma and Max Welling.
\newblock Auto-encoding variational bayes, 2013.
\newblock cite arxiv:1312.6114.

\bibitem{DEN2014}
P.~{Huang}, Y.~{Huang}, W.~{Wang}, and L.~{Wang}.
\newblock Deep embedding network for clustering.
\newblock In {\em ICPR}, pages 1532--1537, Aug 2014.

\bibitem{DMC2017}
Dongdong Chen, Jiancheng Lv, and Yi~Zhang.
\newblock Unsupervised multi-manifold clustering by learning deep
  representation.
\newblock In {\em AAAI Workshops}, volume WS-17 of {\em AAAI Workshops}. AAAI
  Press, 2017.

\bibitem{DSCN2017}
Pan Ji, Tong Zhang, Hongdong Li, Mathieu Salzmann, and Ian Reid.
\newblock Deep subspace clustering networks.
\newblock In {\em NeurIPS}, NIPS'17, pages 23--32, USA, 2017. Curran Associates
  Inc.

\bibitem{Xie2016}
Junyuan Xie, Ross Girshick, and Ali Farhadi.
\newblock Unsupervised deep embedding for clustering analysis.
\newblock In {\em ICML}, ICML'16, pages 478--487. JMLR.org, 2016.

\bibitem{LI2018161}
Fengfu Li, Hong Qiao, and Bo~Zhang.
\newblock Discriminatively boosted image clustering with fully convolutional
  auto-encoders.
\newblock {\em Pattern Recognition}, 83:161 -- 173, 2018.

\bibitem{DCN2016}
Bo~Yang, Xiao Fu, Nicholas~D. Sidiropoulos, and Mingyi Hong.
\newblock Towards k-means-friendly spaces: Simultaneous deep learning and
  clustering.
\newblock In Doina Precup and Yee~Whye Teh, editors, {\em ICML}, volume~70 of
  {\em Proceedings of Machine Learning Research}, pages 3861--3870,
  International Convention Centre, Sydney, Australia, 06--11 Aug 2017. PMLR.

\bibitem{DeepCluster2017}
Kai Tian, Shuigeng Zhou, and Jihong Guan.
\newblock Deepcluster: A general clustering framework based on deep learning.
\newblock In Michelangelo Ceci, Jaakko Hollm{\'e}n, Ljup{\v{c}}o Todorovski,
  Celine Vens, and Sa{\v{s}}o D{\v{z}}eroski, editors, {\em Machine Learning
  and Knowledge Discovery in Databases}, pages 809--825, Cham, 2017. Springer
  International Publishing.

\bibitem{Zhang_2019_CVPR}
Junjian Zhang, Chun-Guang Li, Chong You, Xianbiao Qi, Honggang Zhang, Jun Guo,
  and Zhouchen Lin.
\newblock Self-supervised convolutional subspace clustering network.
\newblock In {\em CVPR}, June 2019.

\bibitem{DEPICT2017}
K.~G. {Dizaji}, A.~{Herandi}, C.~{Deng}, W.~{Cai}, and H.~{Huang}.
\newblock Deep clustering via joint convolutional autoencoder embedding and
  relative entropy minimization.
\newblock In {\em ICCV}, pages 5747--5756, Oct 2017.

\bibitem{VaDE2017}
Zhuxi Jiang, Yin Zheng, Huachun Tan, Bangsheng Tang, and Hanning Zhou.
\newblock Variational deep embedding: An unsupervised and generative approach
  to clustering.
\newblock In {\em IJCAI}, pages 1965--1972, 2017.

\bibitem{GMVAE}
Nat Dilokthanakul, Pedro A.~M. Mediano, Marta Garnelo, Matthew C.~H. Lee, Hugh
  Salimbeni, Kai Arulkumaran, and Murray Shanahan.
\newblock Deep unsupervised clustering with gaussian mixture variational
  autoencoders.
\newblock {\em ArXiv}, abs/1611.02648, 2017.

\bibitem{DASC2018}
Pan Zhou, Yunqing Hou, and Jiashi Feng.
\newblock Deep adversarial subspace clustering.
\newblock In {\em CVPR}, June 2018.

\bibitem{Yang2016Joint}
Jianwei Yang, Devi Parikh, and Dhruv Batra.
\newblock Joint unsupervised learning of deep representations and image
  clusters.
\newblock In {\em CVPR}, 2016.

\bibitem{pmlr-v70-hu17b}
Weihua Hu, Takeru Miyato, Seiya Tokui, Eiichi Matsumoto, and Masashi Sugiyama.
\newblock Learning discrete representations via information maximizing
  self-augmented training.
\newblock In Doina Precup and Yee~Whye Teh, editors, {\em ICML}, volume~70 of
  {\em Proceedings of Machine Learning Research}, pages 1558--1567,
  International Convention Centre, Sydney, Australia, 06--11 Aug 2017. PMLR.

\bibitem{CCNN2017}
C.~{Hsu} and C.~{Lin}.
\newblock Cnn-based joint clustering and representation learning with feature
  drift compensation for large-scale image data.
\newblock {\em IEEE Transactions on Multimedia}, 20(2):421--429, Feb 2018.

\bibitem{inpain2016}
D.~{Pathak}, P.~{Krahenbuhl}, J.~{Donahue}, T.~{Darrell}, and A.~A. {Efros}.
\newblock Context encoders: Feature learning by inpainting.
\newblock In {\em 2016 IEEE Conference on Computer Vision and Pattern
  Recognition (CVPR)}, pages 2536--2544, June 2016.

\bibitem{colorization2016}
Richard Zhang, Phillip Isola, and Alexei~A. Efros.
\newblock Colorful image colorization.
\newblock In {\em ECCV}, volume 9907, pages 649--666. Springer, 2016.

\bibitem{jigsaw2016}
Mehdi Noroozi and Paolo Favaro.
\newblock Unsupervised learning of visual representations by solving jigsaw
  puzzles.
\newblock In Bastian Leibe, Jiri Matas, Nicu Sebe, and Max Welling, editors,
  {\em ECCV}, volume 9910 of {\em Lecture Notes in Computer Science}, pages
  69--84. Springer, 2016.

\bibitem{count2017}
Mehdi Noroozi, Hamed Pirsiavash, and Paolo Favaro.
\newblock Representation learning by learning to count.
\newblock In {\em ICCV}, pages 5899--5907. IEEE Computer Society, 2017.

\bibitem{rotation2018}
Spyros Gidaris, Praveer Singh, and Nikos Komodakis.
\newblock Unsupervised representation learning by predicting image rotations.
\newblock In {\em ICLR}. OpenReview.net, 2018.

\bibitem{vf2018}
Mathilde Caron, Piotr Bojanowski, Armand Joulin, and Matthijs Douze.
\newblock Deep clustering for unsupervised learning of visual features.
\newblock In Vittorio Ferrari, Martial Hebert, Cristian Sminchisescu, and Yair
  Weiss, editors, {\em ECCV}, volume 11218 of {\em Lecture Notes in Computer
  Science}, pages 139--156. Springer, 2018.

\bibitem{NIPS2017}
Ashish Vaswani, Noam Shazeer, Niki Parmar, Jakob Uszkoreit, Llion Jones,
  Aidan~N Gomez, \L~ukasz Kaiser, and Illia Polosukhin.
\newblock Attention is all you need.
\newblock In I.~Guyon, U.~V. Luxburg, S.~Bengio, H.~Wallach, R.~Fergus,
  S.~Vishwanathan, and R.~Garnett, editors, {\em NeurIPS}, pages 5998--6008.
  Curran Associates, Inc., 2017.

\bibitem{Anderson2017Bottom}
Peter Anderson, Xiaodong He, Chris Buehler, Damien Teney, Mark Johnson, Stephen
  Gould, and Lei Zhang.
\newblock Bottom-up and top-down attention for image captioning and visual
  question answering.
\newblock In {\em CVPR}, June 2018.

\bibitem{Zhang}
Han Zhang, Ian~J. Goodfellow, Dimitris~N. Metaxas, and Augustus Odena.
\newblock Self-attention generative adversarial networks.
\newblock {\em ArXiv}, abs/1805.08318, 2018.

\bibitem{Li2018}
Wei Li, Xiatian Zhu, and Shaogang Gong.
\newblock Harmonious attention network for person re-identification.
\newblock In {\em CVPR}, June 2018.

\bibitem{Wang2018}
Qiang Wang, Zhu Teng, Junliang Xing, Jin Gao, Weiming Hu, and Stephen Maybank.
\newblock Learning attentions: Residual attentional siamese network for high
  performance online visual tracking.
\newblock In {\em CVPR}, June 2018.

\bibitem{Liu2018}
Jiang Liu, Chenqiang Gao, Deyu Meng, and Alexander~G. Hauptmann.
\newblock Decidenet: Counting varying density crowds through attention guided
  detection and density estimation.
\newblock In {\em CVPR}, June 2018.

\bibitem{Li2018Tell}
Kunpeng Li, Ziyan Wu, Kuan-Chuan Peng, Jan Ernst, and Yun Fu.
\newblock Tell me where to look: Guided attention inference network.
\newblock In {\em CVPR}, June 2018.

\bibitem{He2018}
Tong He, Zhi Tian, Weilin Huang, Chunhua Shen, Yu~Qiao, and Changming Sun.
\newblock An end-to-end textspotter with explicit alignment and attention.
\newblock In {\em CVPR}, June 2018.

\bibitem{STL2011}
Adam Coates, Andrew~Y. Ng, and Honglak Lee.
\newblock An analysis of single-layer networks in unsupervised feature
  learning.
\newblock In {\em AISTATS}, volume~15 of {\em JMLR Proceedings}, pages
  215--223. JMLR.org, 2011.

\bibitem{CIFAR2009}
Alex Krizhevsky.
\newblock Learning multiple layers of features from tiny images.
\newblock Technical report, University of Toronto, 2009.

\bibitem{hubert1985comparing}
L.~Hubert and P.~Arabie.
\newblock {Comparing partitions}.
\newblock {\em Journal of classification}, 2(1):193--218, 1985.

\bibitem{strehl2002clusterensembles}
Alexander Strehl and Joydeep Ghosh.
\newblock Cluster ensembles -- a knowledge reuse framework for combining
  multiple partitions.
\newblock {\em Journal on Machine Learning Research (JMLR)}, 3:583--617,
  December 2002.

\bibitem{LiD06}
Tao Li and Chris H.~Q. Ding.
\newblock The relationships among various nonnegative matrix factorization
  methods for clustering.
\newblock In {\em ICDM}, pages 362--371. IEEE Computer Society, 2006.

\bibitem{Pasi2006Fast}
P.~{Franti}, O.~{Virmajoki}, and V.~{Hautamaki}.
\newblock Fast agglomerative clustering using a k-nearest neighbor graph.
\newblock {\em IEEE Transactions on Pattern Analysis and Machine Intelligence},
  28(11):1875--1881, Nov 2006.

\bibitem{NMF}
Deng Cai, Xiaofei He, Xuanhui Wang, Hujun Bao, and Jiawei Han.
\newblock Locality preserving nonnegative matrix factorization.
\newblock In {\em IJCAI}, pages 1010--1015, 2009.

\bibitem{Zeiler2010Deconvolutional}
Matthew~D. Zeiler, Dilip Krishnan, Graham~W. Taylor, and Robert Fergus.
\newblock Deconvolutional networks.
\newblock In {\em Computer Vision and Pattern Recognition}, 2010.

\bibitem{SWWAE2015}
Junbo~Jake Zhao, Micha?l Mathieu, Ross Goroshin, and Yann LeCun.
\newblock Stacked what-where auto-encoders.
\newblock {\em CoRR}, abs/1506.02351, 2015.

\bibitem{catgan2016}
Jost~Tobias Springenberg.
\newblock Unsupervised and semi-supervised learning with categorical generative
  adversarial networks.
\newblock In Yoshua Bengio and Yann LeCun, editors, {\em ICLR}, 2016.

\end{thebibliography}

\appendix

\section{Visualization mapping function}
\label{app_vis}
For visualizing the clustering results in Figure 3, the mapping function maps the label feature $l_i \in R^k$ to a two-dimensional vector $z_i \in R^2$. Then, the two-dimensional vectors can be drawn on the figure and rendered by the ground-truth labels.
Specifically, the mapping function is defined as
\begin{equation}
v_i = \left[ \sum_{h=i}^k l_{ih} \sin(\frac{2\pi h}{k}), \sum_{h=i}^k l_{ih} \cos(\frac{2\pi h}{k}) \right] \tag{A1},
\end{equation}
where $\|l_i\|_1 = 1$ due to the probability constraint of label features. For the ImageNet-10, the cluster number $k$ is set to 10.

\section{Proof on trivial solutions of DAC}
\label{app_dac}
In this section, we give the proof on why the DAC is prone to  trivial solutions. Namely, we have following conclusions:

\emph{If the optimal value of Eq. \ref{eq_opt1} is attained, for $\forall i, j, l_i = l_j  \Leftrightarrow r_{ij}=1$ and $r_{ij}=0 \Rightarrow l_i \ne l_j$, but $|\{l_i\}^N_{i=1}| = k$ cannot be guaranteed so that trivial solutions will be obtained.}

\begin{proof}
If the optimal value of Eq. \ref{eq_opt2} is attained, for  $\forall i, j$, we have:
\begin{equation}
l_i \cdot  l_j = \left\{
                                 \begin{array}{cc}
                                 1, \quad & if \  r_{ij}=1, \\
                                 0, \quad & if \ r_{ij}=0,
                                 \end{array}
                                 \right.\tag{A2}
\end{equation}
For $\forall \ i, \|l_i\|_2=1$ and $l_{ih} \ge 0 \  (h=1, \cdots, k)$ are satisfied, where $\|\cdot\|_2$ implies $L_2$-norm of a vector.

First, we verify that $ for \  \forall i, j, l_i = l_j  \Leftrightarrow r_{ij}=1$ and $r_{ij}=0 \Rightarrow l_i \ne l_j$ are satisfied. For arbitrary $l_i, l_j$, if $r_{ij}=1$, we have $l_i \cdot l_j$ = 1 and $\|l_i\|_2=\|l_j\|_2=1$, and we can obtain the following equation:
\begin{equation}
\label{eq_a3}
 \begin{split}
    \|l_i - l_j\|_2^2& = \sum_{h=1}^k(l_{ih}-l_{jh})^2 \\
                                      &=\sum_{k=1}^k(l_{ih}+l_{jh}-2l_{ih}l_{jh}) \\
                                      &=\sum_{h=1}^k l_{ih}^2 + \sum_{h=1}^k l_{jh}^2 - 2\sum_{h=1}^k l_{ih}l_{jh} \\
                                      &=\|l_i\|_2^2 + \|l_j\|_2^2 - 2l_i \cdot l_j \\
                                      &= 1 + 1 - 2\cdot 1 \\
                                      &=0.
\end{split}
\tag{A3}
\end{equation}
 
That is $\bm{l}_i=\bm{l}_j$ is satisfied if $r_{ij}=1$. Similarly, if $f_{ij}=1$, $l_i \ne l_j$ is always satisfied. That is,

\begin{equation}
\label{eq_a4}
\begin{split}
&r_{ij}=1 \Rightarrow l_i = l_j, \\
&r_{ij} = 0 \Rightarrow l_i \ne  l_j, 
\end{split}
\tag{A4}
\end{equation}
Furthermore, due to $\forall \ i, \|l_i\|_2=1$, we have the following proposition if $l_i = l_j$ is satisfied:

\begin{equation}
\label{eq_a5}
r_{ij}= l_i \cdot l_j = l_i \cdot l_i = 1.
\tag{A5}
\end{equation}
According to Eq. (\ref{eq_a4}) and Eq. (\ref{eq_a5}), we have:
\begin{equation}
\label{eq_a6}
r_{ij} = 1 \Leftrightarrow l_i = l_j.
\tag{A6}
\end{equation}
Eq. (\ref{eq_a6}) means that $l_i=l_j$ if and only if $x_i, x_j$ belong to the same clusters. And $r_{ij}=0 \Rightarrow l_i \ne l_j$ implies that $l_i \ne l_j$ if $x_i, x_j$ belong to different clusters. 

However, we assume that $\forall i \ l_i  \in E^k$ that denotes the standard basis of the $k$-dimensional Euclidean space. When $|\{l_i\}^N_{i=1}| < k$, all above equations are true. For example, if $|\{l_i\}^N_{i=1}| = 1$ meaning that all examples belong to the same cluster, while it is also the optimal solution of Eq. \ref{eq_opt1}. For an extreme example, we assume that  $\forall i \ l_i  = [1, 0, \cdots, 0] \in R^k$ , then the estimated $r_{ij}  = 1$ being always true and it is easy  to verify that this solution is an optimal one of Eq. \ref{eq_opt1}.

\end{proof}

\section{Proof of Label Feature Theorem}
\label{app_plft}

In this section, we give formal proof of the Label Feature Theorem:

\begin{lft*}
\label{theorem_p}
If the optimal value of Eq. \ref{eq_opt2} is attained, for $\forall i, j, l_i \in E^k, l_i \neq l_j \Leftrightarrow r_{ij}=0$, $l_i = l_j  \Leftrightarrow r_{ij}=1$, and $|\{l_i\}^N_{i=1}| = k$, where $|\cdot|$ denotes the cardinality of a set.
\end{lft*}

\begin{proof}
If the optimal value of Eq. \ref{eq_opt2} is attained,  we have:

1) For  $\forall i, j$,
\begin{equation}
\label{eq_a7}
y_i \cdot  y_j = \left\{
                                 \begin{array}{cc}
                                 1, \quad & if \  r_{ij}=1, \\
                                 0, \quad & if \ r_{ij}=0,
                                 \end{array}
                                 \right.\tag{A7}
\end{equation}
where for $\forall \ i,  y_i = \frac{l_i}{\|l_i\|_2}, \|l_i\|_2=1$,  $0 \le l_{ih} \le 0 \  (h=1, \cdots, k)$ are satisfied, and $\|\cdot\|_2$ implies $L_2$-norm of a vector so that $\|y_i\|_2 = 1$.

2) For $\forall i$, the value of $l_i \cdot l_i $ is maximized.

\emph{First, we verify that $ for \  \forall i, j, y_i = y_j  \Leftrightarrow r_{ij}=1$  is satisfied.}

For arbitrary $y_i, y_j$, if $r_{ij}=1$, we have $y_i \cdot y_j$ = 1 and $\|y_i\|_2=\|y_j\|_2=1$, and we can obtain the following equation:
\begin{equation}
\label{eq_a8}
 \begin{split}
    \|y_i - y_j\|_2^2& = \sum_{h=1}^k(y_{ih}-y_{jh})^2 \\
                                      &=\sum_{k=1}^k(y_{ih}+y_{jh}-2y_{ih}y_{jh}) \\
                                      &=\sum_{h=1}^k y_{ih}^2 + \sum_{h=1}^k y_{jh}^2 - 2\sum_{h=1}^k y_{ih}y_{jh} \\
                                      &=\|y_i\|_2^2 + \|y_j\|_2^2 -2 y_i \cdot y_j \\
                                      &= 1 + 1 - 2\cdot 1 \\
                                      &=0.
\end{split}
\tag{A8}
\end{equation}
 
That is $\bm{y}_i=\bm{y}_j$ is satisfied if $r_{ij}=1$. That is,
\begin{equation}
\label{eq_a9}
r_{ij} = 1 \Rightarrow l_i = l_j
\tag{A9}
\end{equation}

Furthermore, due to $\forall i, \|y_i\|_2=1$, we have the following proposition if $y_i = y_j$ is satisfied:
\begin{equation}
\label{eq_a10}
r_{ij} = y_i \cdot y_j = y_i \cdot y_i = 1.
\tag{A10}
\end{equation}
According to Eq. \ref{eq_a9} and Eq. \ref{eq_a10}, we have 

\begin{equation}
\label{eq_a11}
r_{ij} = 1 \Leftrightarrow y_i = y_j
\tag{A11}
\end{equation}

 In the probability constraint, $\forall i,j, \sum_h l_{ih} = \sum_h l_{jh} = 1$ is always true, then we have:
\begin{equation}
\begin{split}
\bm{y}_{i} = \bm{y}_{j} & \Leftrightarrow \forall h, \frac{l_{ih}}{\|l_i\|_2} = \frac{l_{jh}}{\|l_j\|_2} \\
&\Leftrightarrow \forall h, l_{ih} = l_{jh} \frac{\|l_i\|_2}{\|l_j\|_2}\\
&\Leftrightarrow \sum_h l_{jh}  \frac{\|l_i\|_2}{\|l_j\|_2} = \sum_h l_{jh} \\
& \Leftrightarrow  \left( \frac{\|l_i\|_2}{\|l_j\|_2} - 1 \right) \sum_h l_{jh} = 0 \\
& \Leftrightarrow  \frac{\|l_i\|_2}{\|l_j\|_2} = 1 \\
& \Leftrightarrow \forall h, l_{ih} = l_{jh} \\
& \Leftrightarrow l_i = l_j
\end{split}
\tag{A12}
\end{equation}
According to Eq. \ref{eq_a11} and Eq. \ref{eq_a10}, we have $r_{ij} = 1 \Leftrightarrow l_i = l_j$, meaning that $l_i = l_j$ if and only if $x_i, x_j$ belong to the same cluster.

\emph{Second, we verify that $\forall i, l_i \in E^k$ that denotes the standard basis of the $k$-dimensional Euclidean space, and $r_{ij} = 0 \Leftrightarrow l_i \ne l_j$. }

Due to $\forall i,h, 0 \le l_{ih} \le 1$, then $ 0 \le l^2_{lh} \le l_{lh} \Rightarrow l_i \cdot l_i = \sum_h l_{ih}^2 \le \sum_h l_{ih} = 1$. Therefore, if the optimal solution of Eq. \ref{eq_opt2} is attained, the value of $l_i \cdot l_i$ is maximized, i.e., $l_i \cdot l_i = 1$. Then, we verify the  proposition that $l_i \cdot l_i = 1$ is satisfied if and only if $l_i$ is the one-hot vector, i.e, $ \forall i, \exists m, l_{im} = 1, \forall h \ne m, l_{ih} = 0$. 

Considering that $\forall i, \sum_h l_{ih} = 1$ and $\forall i, h, 0 \le l_{ih} \le 1$, its opposite proposition is that $\forall i, h, 0 \le l_{ih} < 1$. We assume the opposite proposition is correct, then we have $\forall i, h, l_{ih}^2 < l_{ih} \Rightarrow \forall i, l_i \cdot l_i  =  \sum_h l_{ih}^2 < \sum_h l_{ih} = 1$. It is contradictory to that $l_i \cdot l_i = 1$, so the original proposition is true.

As $\forall i, l_i$  is the one-hot vector, we can easily verify that $r_{ij} = 0 \Rightarrow l_i \ne l_j$ and $l_i \ne l_j \Rightarrow r_{ij} = 0$, thus $r_{ij} = 0 \Rightarrow l_i \ne l_j$.

\emph{Finally, we verify the proposition that $|\{l_i\}^N_{i=1}| = k$ is satisfied.}

We have the nonempty cluster constraint that $\forall h, p_h = \frac{1}{N}\sum_{i=1}^N l_{ih} $ > 0, so that $\forall h, \exists i, l_{ih} > 0$. As we have already proved that $\forall i, l_i \in E^k$, therefore, $|\{l_i\}_{i=1}^N|=|E^k|=k$.

\end{proof}

\section{Architecture details for all experiments}
\label{app_arc}

In this section, we describe the architecture details for each experiment. The proposed GATCluster consists of three components, i.e., the image feature module, the label feature module, and the attention module. The architectures of the image feature module for different experiments are summarized in Table \ref{table_base}. All of them are implemented by the VGG-style convolutional neural networks with batch normalization. On different datasets, the architectures slightly differ from each other in the layer number, kernel size of the first layer and padding value. To evaluate the effectiveness of attention map resolution, we simply stacked extra convolutional layers to the last block and the feature map resolution is reduced due to the valid convolution, i.e., $3\times3$ kernel size with zero-padding. Similarly, to further increase the input image size (i.e., $160 \times 160$ and $192\times192$) and keep the attention-map-size unchanged (i.e., $6\times6$), we also stacked extra convolutional layers to the last block, which is similar to that of ImageNet-10-128. Without too much redundancy, the architecture details for the large input-size of $160 \times 160$ and $192\times192$ are not shown in the Table \ref{table_base}. For the label feature module and attention module, all experiments share the same network configuration, as shown in Table \ref{table_label} and \ref{table_attention}.

In this work, we mainly focus on the learning framework and the corresponding theory of the deep unsupervised clustering. We did not study the network architectures comprehensively. To further improve the clustering performance, we believe it is a promising way to search an effective architecture for deep clustering, especially, in an automatic manner. 

\section{Definitions of evaluation metrics}
\label{sec_def}
In this section, we introduce the definition of the evaluation metrics as following.

\textbf{\emph{Accuracy (ACC)}}. Given the predicted label $q_i$ and the ground-truth label $r_i$ of each sample, a mapping function $O$ is first obtained by Hungarian algorithm 
to find the correspondence between $q_i$ and $r_i$. If $r_i = O(q_i)$, the sample is correctly predicted, or otherwise, the sample is falsely predicted. Then, the ACC is calculated as:
\begin{equation}
ACC = \frac{\sum_{i=1}^{N}\delta(r_i, O(q_i))}{N} \tag{A13}
\end{equation}
where
\begin{equation}
\delta(a, b) = \left\{
\begin{array}{ccc}
1, &    & a=b, \\
0, &    & otherwise,
\tag{A14}
\end{array}
\right.
\end{equation}
and N is the total number of samples.

\textbf{\emph{Normalized Mutual Information (NMI)}}. Given the predicted partition $U=\{U_1, \cdots, U_R\}$ and the ground-truth partition $V=\{V_1,\cdots,V_C\}$, the NMI is defined by:
\begin{equation}
NMI = \frac{\sum_{i=1}^{R}\sum_{j=1}^{C}|U_i\cap V_j|\log \frac{N|U_i\cap V_j|}{|U_i||V_j|}}{\sqrt{(\sum_{i=1}^{R}|U_i|\log \frac{|U_i|}{N})(\sum_{j=1}^{C}|V_j|\log \frac{|V_j|}{N})}} \tag{A15}
\end{equation}
where $N$ is the total number of samples, $R$ and $C$ mean the partition number, $|U_i|$ presents the sample number of the subset $U_i$.

\textbf{\emph{Adjusted Rand Index (ARI)}}. Given the predicted partition $U=\{U_1, \cdots, U_R\}$ and the ground-truth partition $V=\{V_1,\cdots,V_C\}$, the information on class overlap between the two partitions $U$ and $V$ can be obtained as shown in the Table \ref{table_ari},
\begin{table}[h]
\footnotesize
\renewcommand{\arraystretch}{1.2}
\renewcommand\tabcolsep{2pt}
\renewcommand{\thetable}{S\arabic{table}}
\caption{Contingency table.}
\label{table_ari}
\centering
\begin{tabular}{c|cccc|c}

Class & $V_1$ & $V_2$ & $\cdots$ &$V_C$ & Sums \\
\hline
$U_1$&$n_{11}$&$n_{12}$&$\cdots$&$n_{1C}$&$n_{1\cdot}$\\
$U_2$&$n_{21}$&$n_{22}$&$\cdots$&$n_{2C}$&$n_{2\cdot}$\\
$\cdot$&$\cdot$&$\cdot$&$\cdot$&$\cdot$&$\cdot$\\
$\cdot$&$\cdot$&$\cdot$&$\cdot$&$\cdot$&$\cdot$\\
$\cdot$&$\cdot$&$\cdot$&$\cdot$&$\cdot$&$\cdot$\\
$U_R$&$n_{R1}$&$n_{R2}$&$\cdots$&$n_{RC}$&$n_{R\cdot}$\\
\hline
Sums&$n_{\cdot 1}$&$n_{\cdot 2}$&$\cdots$&$n_{\cdot C}$&$n_{\cdot \cdot}=N$\\
\end{tabular}
\end{table}
where $n_{ij}$ denotes the number of sample of objects that are common to subset $U_i$ and $V_j$, $n_{i\cdot}$ and $n_{\cdot j}$ denotes the number of samples in the subset $U_i$ and $U_j$ respectively, $R$ and $C$ mean the partition number, and $N$ is the total number of samples. Then, the ARI is defined as:
\begin{equation}
ARI = \frac{\sum_{ij}\left(_{2}^{n_{ij}}\right) - \left[\sum_i\left(_{2}^{n_{i\cdot}}\right)  \sum_j\left(_{2}^{n_{\cdot j}}\right)  \right] / \left(_{2}^{n}\right)}{\frac{1}{2} \left[\sum_i\left(_{2}^{n_{i\cdot}}\right) +  \sum_j\left(_{2}^{n_{\cdot j}}\right)  \right] - \left[\sum_i\left(_{2}^{n_{i\cdot}}\right)  \sum_j\left(_{2}^{n_{\cdot j}}\right)  \right] / \left(_{2}^{n}\right)}, \tag{A16}
\end{equation}
where the $\left(_{2}^{n}\right)$ is the binomial coefficient.

\begin{table*}[ht]
\scriptsize
\renewcommand{\arraystretch}{1.2}
\renewcommand\tabcolsep{1.2pt}
\renewcommand{\thetable}{S\arabic{table}}
\caption{Architecture details of image feature module for different experiments. The number format is $k_h$$\times$$k_w$-$s$-$p$-$c$, where $k_h$, $k_w$, $s$, $p$, and $c$ represent the kernel height, kernel width, stride, padding, and output channel number, respectively. Conv BN ReLU represent the convolution layer followed by a batch normalization layer with a ReLU activation. For simplicity, Conv-256 means 3$\times$3-1-0-0-256 Conv BN ReLU, and Conv-k means 1$\times$1-1-0-$k$ Conv BN ReLU. The MaxPooling has the similar number format except that it has no output channel number. $k$ is the cluster number.}
\label{table_base}
\centering
\begin{tabular}{|c|c|c|c|c|c|c|c|}

\hline
STL10/ImageNet-10 & ImageNet-Dog & Cifar10/Cifar100-20& ImageNet-Dog-128& \multicolumn{4}{c|}{ImageNet-10-128-Att-10/8/4/2}  \\
\hline \hline
3$\times$3-1-0-64 Conv BN ReLU  & 5$\times$5-1-1-64 Conv BN ReLU & 3$\times$3-1-1-64 Conv BN ReLU & 7$\times$7-1-1-64 Conv BN ReLU &\multicolumn{4}{c|}{3$\times$3-1-0-64 Conv BN ReLU}\\
3$\times$3-1-0-64 Conv BN ReLU  & 3$\times$3-1-0-64 Conv BN ReLU & 3$\times$3-1-1-64 Conv BN ReLU & 3$\times$3-1-0-64 Conv BN ReLU &\multicolumn{4}{c|}{3$\times$3-1-0-64 Conv BN ReLU}\\
3$\times$3-1-0-64 Conv BN ReLU  & 3$\times$3-1-0-64 Conv BN ReLU & 3$\times$3-1-1-64 Conv BN ReLU & 3$\times$3-1-0-64 Conv BN ReLU &\multicolumn{4}{c|}{3$\times$3-1-0-64 Conv BN ReLU}\\
2$\times$2-2-0 MaxPooling  & 2$\times$2-2-0 MaxPooling & 2$\times$2-2-0 MaxPooling & 2$\times$2-2-0 MaxPooling &\multicolumn{4}{c|}{2$\times$2-2-0 MaxPooling}\\
\hline
3$\times$3-1-0-128 Conv BN ReLU  & 3$\times$3-1-0-128 Conv BN ReLU & 3$\times$3-1-0-128 Conv BN ReLU & 3$\times$3-1-0-128 Conv BN ReLU & \multicolumn{4}{c|}{3$\times$3-1-0-128 Conv BN ReLU} \\
3$\times$3-1-0-128 Conv BN ReLU  & 3$\times$3-1-0-128 Conv BN ReLU & 3$\times$3-1-0-128 Conv BN ReLU & 3$\times$3-1-0-128 Conv BN ReLU  &\multicolumn{4}{c|}{3$\times$3-1-0-128 Conv BN ReLU} \\
3$\times$3-1-0-128 Conv BN ReLU  & 3$\times$3-1-0-128 Conv BN ReLU & 3$\times$3-1-0-128 Conv BN ReLU & 3$\times$3-1-0-128 Conv BN ReLU &\multicolumn{4}{c|}{3$\times$3-1-0-128 Conv BN ReLU} \\
2$\times$2-2-0 MaxPooling  &2$\times$2-2-0 MaxPooling& 2$\times$2-2-0 MaxPooling & 2$\times$2-2-0 MaxPooling&\multicolumn{4}{c|}{2$\times$2-2-0 MaxPooling}\\
\hline
Conv-256  & Conv-256 & Conv-$k$& Conv-256 &  \multicolumn{4}{c|}{Conv-256}\\
Conv-256  & Conv-256 & & Conv-256 &   \multicolumn{4}{c|}{Conv-256}\\
Conv-256  & Conv-256 & &Conv-256 &   \multicolumn{4}{c|}{Conv-256}\\
2$\times$2-2-0 MaxPooling  &2$\times$2-2-0 MaxPooling  & & 2$\times$2-2-0 MaxPooling & \multicolumn{4}{c|}{2$\times$2-2-0 MaxPooling} \\
\hline
Conv-$k$  &Conv-256& & Conv-256&Conv-$k$ &Conv-256&Conv-256&Conv-256\\
                                                            & Conv-$k$ & & Conv-256&                 &Conv-$k$&Conv-256&Conv-256 \\
                                                            &                  & & Conv-256&                 &                 &Conv-256&Conv-256\\
                                                            &                  & & Conv-$k$&                  &                 &Conv-$k$&Conv-256\\
                                                            &                  & &                &                  &                 &                &Conv-$k$\\
\hline

\end{tabular}
\end{table*}

\begin{table}[h]
\footnotesize
\renewcommand{\arraystretch}{1.2}
\renewcommand\tabcolsep{2pt}
\renewcommand{\thetable}{S\arabic{table}}
\caption{Architecture of label feature module.}
\label{table_label}
\centering
\begin{tabular}{|c|}

\hline
STL10/ImageNet-10/ImageNet-Dog/Cifar10/Cifar100-20 \\
\hline \hline

1$\times$1-1-0-64 Conv BN ReLU \\
GlobalPooling \\
Linear(k, k) ReLU \\
Linear(k, k) ReLU \\
Linear(k, k) Softmax \\
\hline

\end{tabular}
\end{table}

\begin{table}[ht]
\footnotesize
\renewcommand{\arraystretch}{1.2}
\renewcommand\tabcolsep{2pt}
\renewcommand{\thetable}{S\arabic{table}}
\caption{Architecture of attention module.}
\label{table_attention}
\centering
\begin{tabular}{|c|}

\hline
STL10/ImageNet-10/ImageNet-Dog/Cifar10/Cifar100-20 \\
\hline \hline

Linear(H*W, 3) ReLU \\
Gaussian-kernel \\
Channel-Multiplication \\
GlobalPooling \\
Linear(k, k) ReLU \\
Linear(k, k) ReLU \\
Linear(k, k) Softmax \\
\hline

\end{tabular}
\end{table}

\section{More implementation details}
\label{sec_exp}

The proposed GATCluster was implemented based on Pytorch. 
By adjusting the sub-batch size $m_1$, our algorithm can fit for different GPUs with both large and relative small memory. For example, we conducted the GATCluster on serval types of devices, including the GTX 1080 with 8G memory, the TITAN Xp and TITAN V with 12G memory, and the Tesla V100 with 32G memory. In this work, the models were trained and tested on a single GPU.

\end{document}